\documentclass[letterpaper, 10 pt, conference]{ieeeconf}
\pdfoutput=1
\usepackage{amsmath}
\usepackage{amssymb}

\usepackage{subcaption}
\usepackage{float}

\IEEEoverridecommandlockouts                             
\usepackage{graphicx}
\usepackage[dvipsnames]{xcolor}

\definecolor{flatgreen}{HTML}{2ECC71}
\definecolor{flatpurple}{HTML}{9B59B6}
\definecolor{deepskyblue}{HTML}{00bfff}

\newcommand{\RRR}{{\mathbb{R}}}

\title{Deep Reinforcement Learning using Cyclical Learning Rates}

\author{Ralf Gulde*, Marc Tuscher*, Akos Csiszar,  Oliver Riedel and Alexander Verl
\thanks{R. Gulde, M. Tuscher, A. Csiszar, O. Riedel and A. Verl
are   with   the   Institute   of   Control   Engineering   of   Machine   Tools   and   
Manufacturing  Units,  University  of  Stuttgart,  70174  Stuttgart,  Germany  
(phone:      0049-711-685-82505;      fax:      0049-711-685-72505;      e-mail:      
ralf.gulde@isw.uni-stuttgart.de). The Research is supported by the Graduate School of Excellence advanced Manufacturing Engineering and by the German Research Foundation (DFG).
* These authors contributed equally.}}

\begin{document}

\maketitle
\thispagestyle{empty}
\pagestyle{empty}

\begin{abstract}

Deep Reinforcement Learning (DRL) methods often rely on the meticulous tuning of hyperparameters to successfully resolve problems. 
One of the most influential parameters in optimization procedures based on stochastic gradient descent (SGD) is the learning rate.
We investigate cyclical learning and propose a method for defining a general cyclical learning rate for various DRL problems.
In this paper we present a method for cyclical learning applied to complex DRL problems.
Our experiments show that, utilizing cyclical learning achieves similar or even better results than highly tuned fixed learning rates. 
This paper presents the first application of cyclical learning rates in DRL settings and is a step towards overcoming manual hyperparameter tuning.

\end{abstract}

\section{Introduction}

Driven by the rapid increase of the amount of available data and computational resources, models and algorithms for deep neural networks have undergone remarkable developments and are state of the art in addressing fundamental tasks ranging from computer vision problems like image classification~\cite{he2016deep}, scene segmentation~\cite{he2017mask}, face recognition~\cite{schroff2015facenet} to natural language processing.

In Machine Learning (ML) hyperparameter tuning is the problem of selecting a set of hyperparameters for an optimal learning strategy. A hyperparameter is a parameter value that is used to control the learning process. The learning speed $\eta$ is an essential hyperparameter and controls the rate of updating the model. In particular, $\eta$ controls the amount of error assigned to update the weights of the model. 
The use of deep neural networks has brought significant progress in solving challenging problems in various fields using Deep Reinforcement Learning (DRL). 
Reproducing existing work and accurately evaluating improvements offered by novel methods is vital to maintain this progress.

DRL problems vary from supervised deep learning problems in one important aspect: the distribution from which the data is taken is non-stationary.
Transferring learning rate techniques from deep learning to deep reinforcement learning is therefore not a trivial task. A common method for determining the learning rate in supervised deep learning approaches is the learning rate decay.
However, due to the non-stationarity of the RL problem, training with linearly decreasing learning rates is inferior, and training should be done at different learning rates~\cite{schaulNoMorePesky2013}.








The main contributions of this work are:
\begin{itemize}
    \item We apply learning rate cycling to DRL problems by training an agent with the PP02 \cite{schulman2017proximal} algorithm. 
    \item We reduce the necessity of a manual learning rate tuning for particular RL environments by applying our general cyclical learning rate for a variety of complex RL problems.
    \item Our experiments show that cyclical learning rates achieve similar or even better performance in various environments compared to fixed learning rates. 
\end{itemize}

The paper is organized as follows: In section~\ref{sec:related} we provide a brief overview of the related work on the learning rate cycling, followed by a brief recapitulation of the theoretical foundations of our method in section~\ref{sec:background}.
Section~\ref{sec:method} introduces the proposed method, section~\ref{sec:experiments} describes our experiments, and section~\ref{sec:ablation} illustrates the critical parameters of our method.
Concluding remarks are given in section\ref{sec:conclusion}.

\section{Related Work}\label{sec:related}

Vanilla gradient descent methods can be made more reliable via line search~\cite{dennis1996numerical}.
Line search relies on computing the full loss on the dataset to find a good learning rate.
However, computing the full loss and therefore the full first derivative is often computationally expensive in settings where stochastic gradient descent (SGD) methods are used.

Schaul~et.~al~\cite{schaulNoMorePesky2013} propose an algorithm to compute optimal learning rates for SGD on non-stationary problems.
This method relies on expectations of the gradient and the square norm of the gradient.

Smith proposes cyclical learning rates for supervised deep learning in~\cite{smithCyclicalLearningRates2017}.
Step sizes are estimated based on the size of an epoch.
The base and maximum learning rate are determined using a method for learning rate identification in which a model is trained for an epoch with linearly increasing learning rates.
Experiments show that cyclical learning rates lead to significant performance improvements.
Learning rate cycling is also combined with a momentum that increases and decreases counteractingly to the learning rate~\cite{smith2018disciplined}. This improves the stability of the training process and increases the training speed, which is further improved by setting a very high maximum learning rate.
However, since high learning rates impose regularization, the number of other regularization techniques must be reduced~\cite{smith2019super}.

To the best of the authors' knowledge, there is no known approach that employs cyclical learning rates for DRL problems. The introduction of a general cyclical learning rate results in the acknowledgement of the non-stationarity of the DRL problem and the overcoming of manual hyperparameter tuning.

\section{Background}\label{sec:background}

This section covers the theoretical foundations for our experiments.

\subsection{Reinforcement Learning and Policy Optimization}

We define the problem of reinforcement learning (RL) and introduce the notation we use throughout the paper.
In this paper we consider discounted Markov decision processes (MDP) with a finite horizon.
At each time step $t$ the RL agent observes the current state $s_t \in S$, performs an action $a_t \in A$ and then receives a reward $r_{t+1} \in \RRR$.
After that the resulting state $s_{t+1}$ will be observed, determined by the unknown dynamics of the environment $p(s_{t+1}| a_t, s_t)$. An episode has a pre-defined length $T$ time steps.
The goal of the agent is to find a parameter $\theta$ of a policy ${\pi_{\theta} (a|s)}$ that maximizes the expected cummulated reward $J$ over a trajectory
\begin{equation} \label{bell}
J(\pi_{\theta}) = \mathbb{E}_{\tau\sim{\pi_{\theta}}} \left[ \sum_{t=0}^T \pi(a_t | s_t) \sum_{k=t}^T \gamma^{k-t} r_{k+1} \right],
\end{equation}
where $\gamma \in [0, 1]$ is the discount factor. 

RL methods solve an MDP by interacting with the system and accumulating the reward achieved.
We consider several model-free policy gradient algorithms with open source implementations that are common in the literature, e.g., Soft Actor Critic ~\cite{haarnoja2018soft}, Deep Deterministic Policy Gradient (DDPG)~\cite{silver2014deterministic}, and Proximal Policy Optimization (PPO)~\cite{schulman2017proximal}. Generally, PPO maximizes (\ref{bell}) using a robust version of the policy gradient theorem
\begin{equation*}\label{pg}
\nabla_{\theta} J(\pi_{\theta}) = \mathbb{E}_{\tau\sim{\pi_{\theta}}} \bigg[ \sum_{t=0}^T \nabla_{\theta} \log \pi(a_t | s_t) \sum_{k=t}^T \gamma^{k-t} r_{k+1} \bigg]
\end{equation*}
by performing gradient ascent steps
\begin{equation}
    \theta_{k+1} = \theta_{k} + \alpha \nabla_{\theta}J(\pi_{\theta}).
\end{equation}

\subsection{Non-Stationarity of RL Problems}
Deep reinforcement learning and supervised deep learning differ in an important aspect: while the general optimization methods and models can be quite similar, the data in deep RL methods is sampled from a non-stationary distribution.
That is, the data is sampled from an environment according to a continuously changing policy $\pi_{\theta}$.
Every time we update our parameters $\theta$, the policy $\pi_{\theta}$ changes, and since every trajectory $\tau$ is sampled while following $\pi_{\theta}$
$$P(\tau|\pi_{\theta}) = \rho_0 (s_0) \prod_{t=0}^{T-1} P(s_{t+1} | s_t, a_t) \pi_{\theta}(a_t | s_t)$$
our data distribution changes with a change in policy.
Off-policy RL methods often save trajectories $\tau$ in a replay buffer, while on-policy methods use the sampled trajectories only for a single policy update.

\section{Cyclical Learning Rates for DRL}\label{sec:method}
\begin{figure}[t]
  \centering
  \includegraphics[width=.8\linewidth]{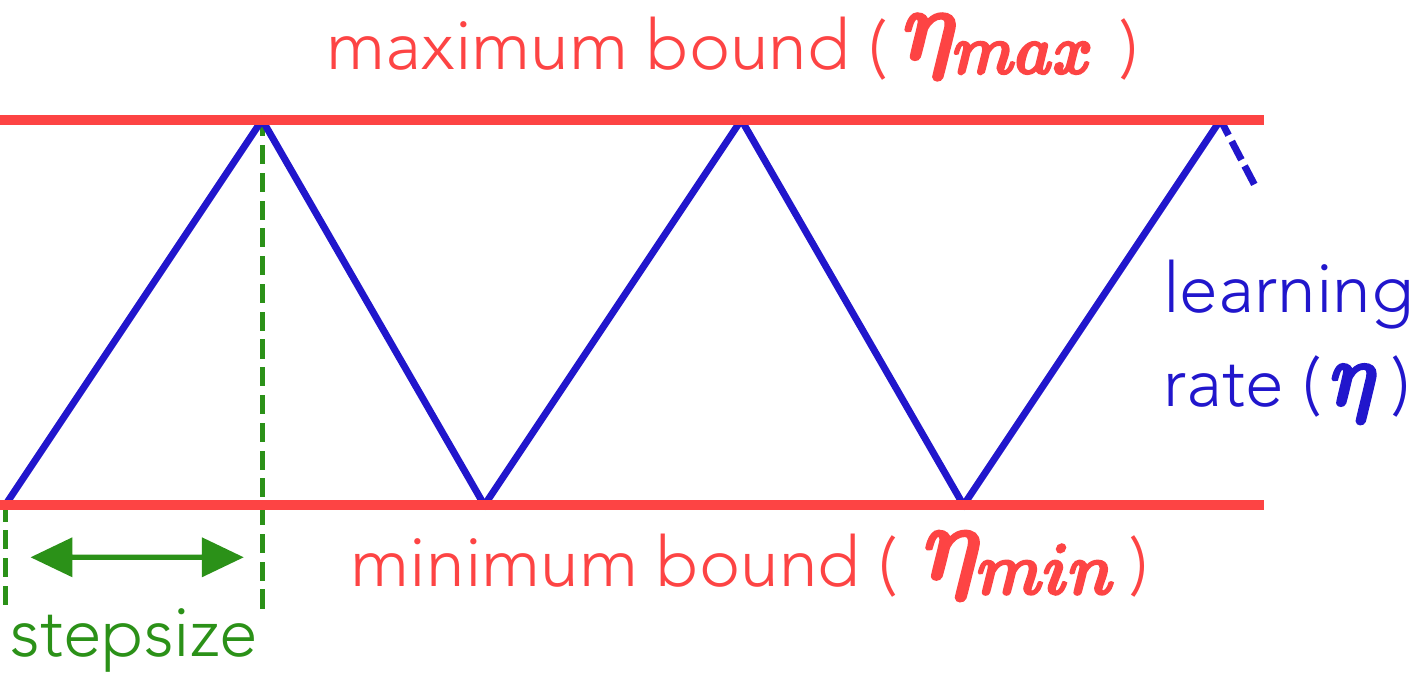}
  \caption[Learning Rate Cycle]{Cyclical learning rate with maximum and minimum bound and stepsize~\cite{smithCyclicalLearningRates2017}.}
  \label{fig:method:cycle}
\end{figure}

\begin{figure}[b]
  \centering
  \includegraphics[width=\linewidth]{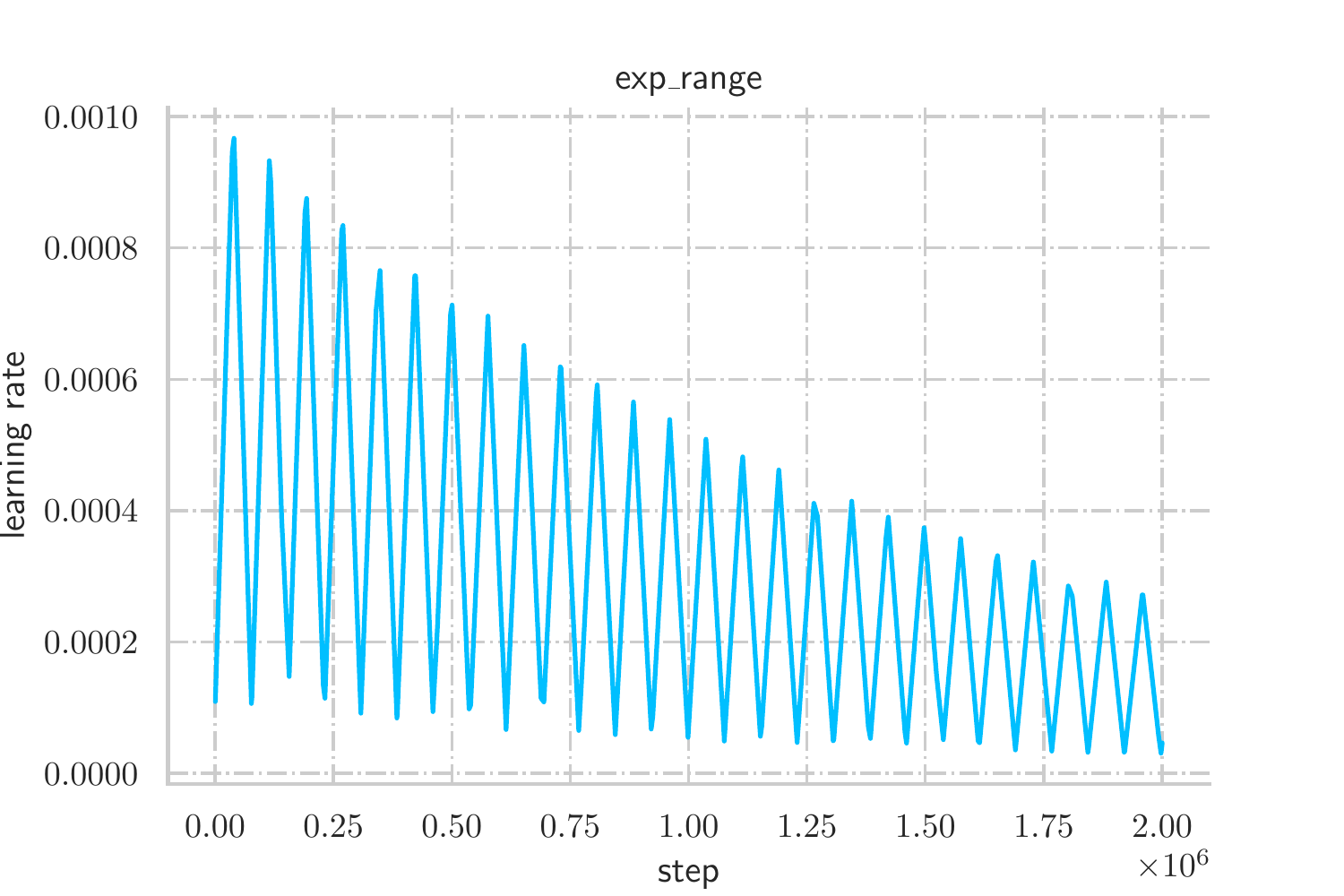}
  \caption[Learning Rate Cycle]{Exponential decay of $\eta_{max}$ and $\eta_{min}$ using the \textit{exp\_range} method.}\label{fig:method:exprange}
\end{figure}

We apply two kinds of cyclical learning rate policies to the DRL problem, the \textit{triangular-} and the \textit{exp\_range-} policy.
Herein, we comply to the terminology used in the original paper introducing cyclical learning rates~\cite{smithCyclicalLearningRates2017}.
The \textit{triangular} policy linearly increases the learning rate for \textit{stepsize} number of iterations and then linearly decreases for \textit{stepsize} iterations.
Increments and decrements occur between the maximum bound ($\eta_{max}$) and the minimum bound ($\eta_{min}$).
This approach is depicted in figure~\ref{fig:method:cycle}.
The \textit{exp\_range} policy inherits the properties of the \textit{triangular} policy, while also decaying the maximum $\eta_{max}$ and the minimum learning rate $\eta_{min}$ over time.
Before each cycle, we compute new learning rate bounds $\eta_{min}$ and $\eta_{max}$ by
\begin{equation*}
\begin{split}
    \eta_{min} &= \eta_{min, 0} \cdot \lambda^{k_{cycle}}\\
    \eta_{max} &= \eta_{max, 0} \cdot \lambda^{k_{cycle}},
\end{split}
\end{equation*}
where $\lambda$ is an exponential decay factor and $k_{cycle}$ is the current cycle count.
The learning rate over timestep using this method is depicted in figure~\ref{fig:method:exprange}, starting with $\eta_{max, 0}=0.001,~~\eta_{min, 0}=0.0001$ and an exponential decay factor $\lambda = 0.99$.
\begin{figure*}[!t]
\begin{subfigure}[b]{.2\textwidth}
  \centering
      \includegraphics[width=\linewidth]{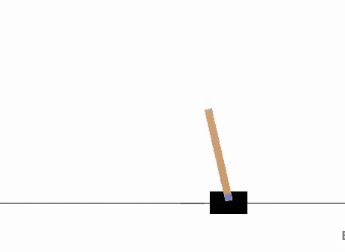}
  \caption{CartPole-v0}\label{fig:envs:cartpole}
\end{subfigure}
\hfill
\begin{subfigure}[b]{.2\textwidth}
  \centering
  \includegraphics[width=\linewidth]{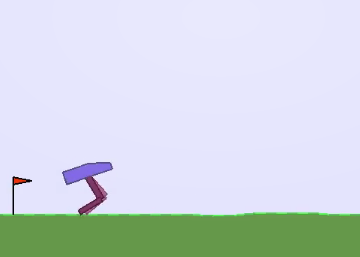}  
  \caption{BipedalWalker-v3}\label{fig:envs:bipedal}
\end{subfigure}
\hfill
\begin{subfigure}[b]{.2\textwidth}
  \centering
  \includegraphics[trim=0 0 0 1.3cm,clip,width=\linewidth]{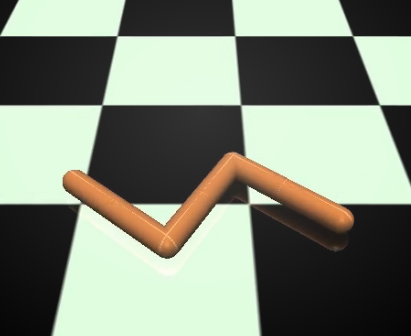}  
  \caption{Swimmer-v2}\label{fig:envs:swimmer}
\end{subfigure}
\hfill
\begin{subfigure}[b]{.2\textwidth}
  \centering
    \includegraphics[trim=0 0 0 1.8cm,clip,width=\linewidth]{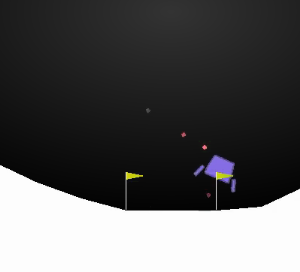}  
  \caption{LunarLander-v3}\label{fig:envs:lunar}
\end{subfigure}
\caption{RL benchmarking environments used in our experiments.}\label{fig:envs}
\vspace{-15pt}
\end{figure*}

Since no notion of training epochs is present in deep RL, epochs cannot be used to provide a good estimate for the step size.
We use a fixed step size for all training runs.
Further, we use fixed $\eta_{max}$ and $\eta_{min}$, since determining reasonable  learning rates for a non-stationary problem is non-trivial, as depicted in section~\ref{sec:ablation}.

To increase the stability of the training process at high learning rates, we increase and decrease the momentum of the optimizer anti-proportional to the learning rate.
That is, when the learning rate increases, the momentum decreases and vice versa.

\section{Experiments}\label{sec:experiments}
The Experiments are performed on various reinforcement learning environments contained in \textit{openai/gym}.
In each experiment, three reinforcement learning agents are trained using the PPO2 algorithm (on the same environment) with the same random seed.
The optimal hyperparameters for each environment are obtained from \textit{rl-zoo}~\cite{rl-zoo}.
Two agents are trained using the \textit{triangular} and the \textit{exp\_range} policy, respectively, the remaining agent is trained with the optimal fixed learning rate for this environment.
We compare the episode reward over time steps of all agents.

In a variety of environments, the \textit{triangular} learning rate policy performs similar or better than the optimal fixed learning rate for the particular environment.
Notably, we do not tune $\eta_{min}$, $\eta_{max}$ nor the step size $s$ for any of our environments, but perform all experiments with a general setting for learning rate cycling. 
These settings are $\eta_{max}=0.01,~~\eta_{min}=0.0001,~~s=2000$.
For the \textit{exp\_range} policy additionally use a decay factor $\lambda = 0.99$.
Further, as described in section~\ref{sec:method} we cycle the momentum of the optimizer between $1.0$ and $0.8$, anti-proportional to the learning rate.

\begin{figure*}[!t]
\begin{subfigure}{.49\textwidth}
  \centering
      \includegraphics[width=\linewidth]{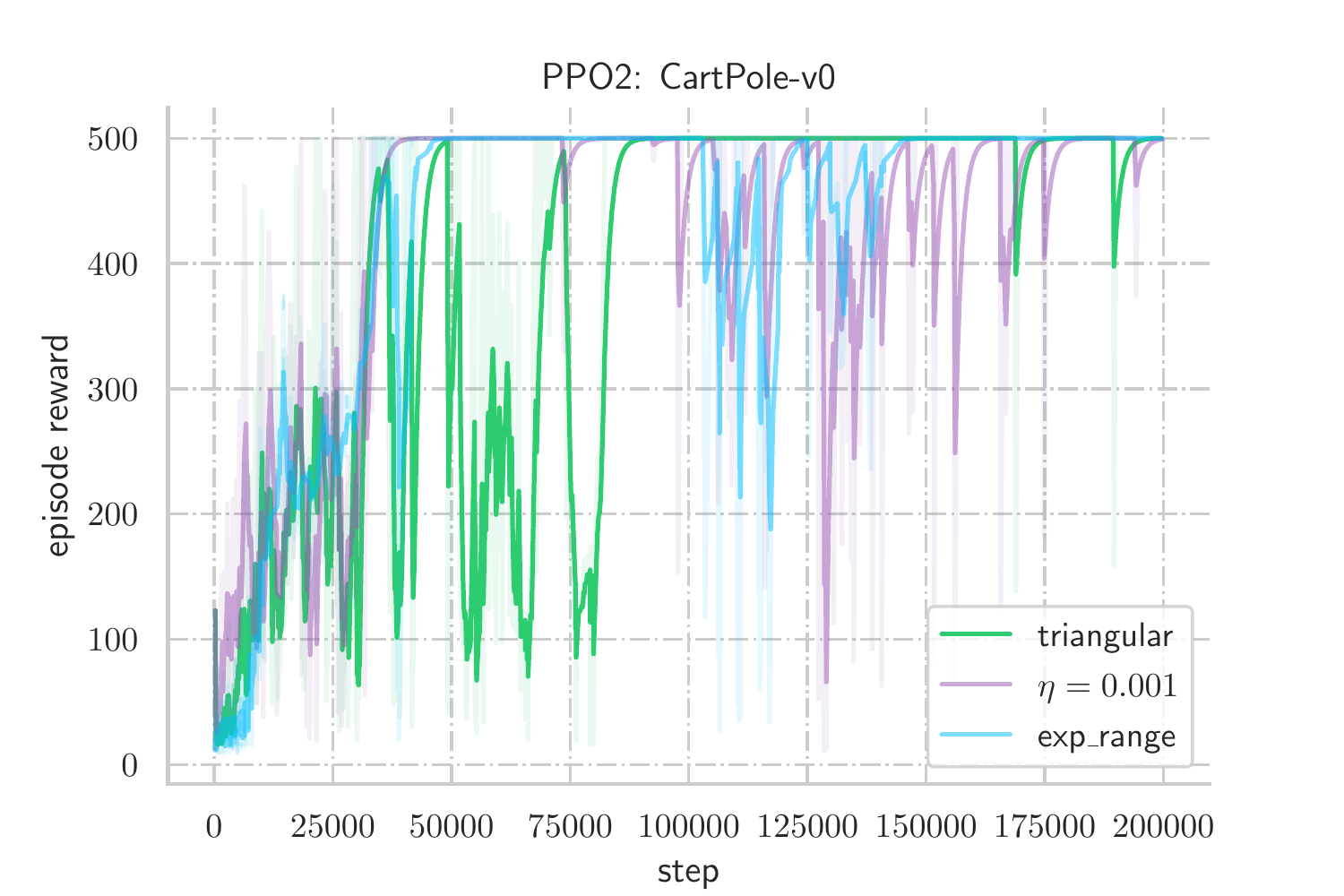}
  \caption{}\label{fig:experiments:cartpole}
\end{subfigure}
\hfill
\begin{subfigure}{.49\textwidth}
  \centering
  \includegraphics[width=\linewidth]{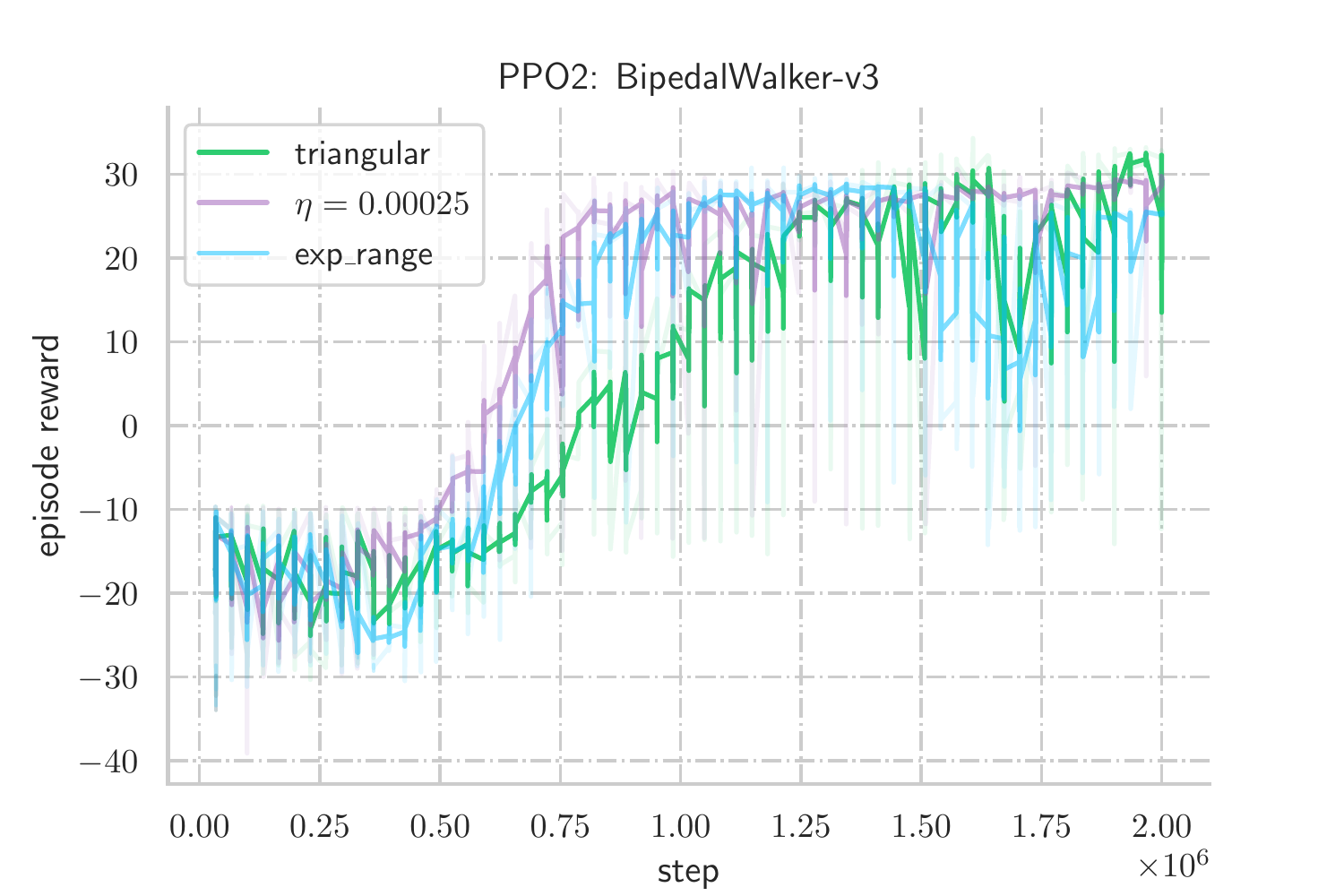}  
  \caption{}\label{fig:experiments:bipedal}
\end{subfigure}
\vskip\baselineskip
\vspace{-12pt}
\hfill
\begin{subfigure}{.49\textwidth}
  \centering
  \includegraphics[width=\linewidth]{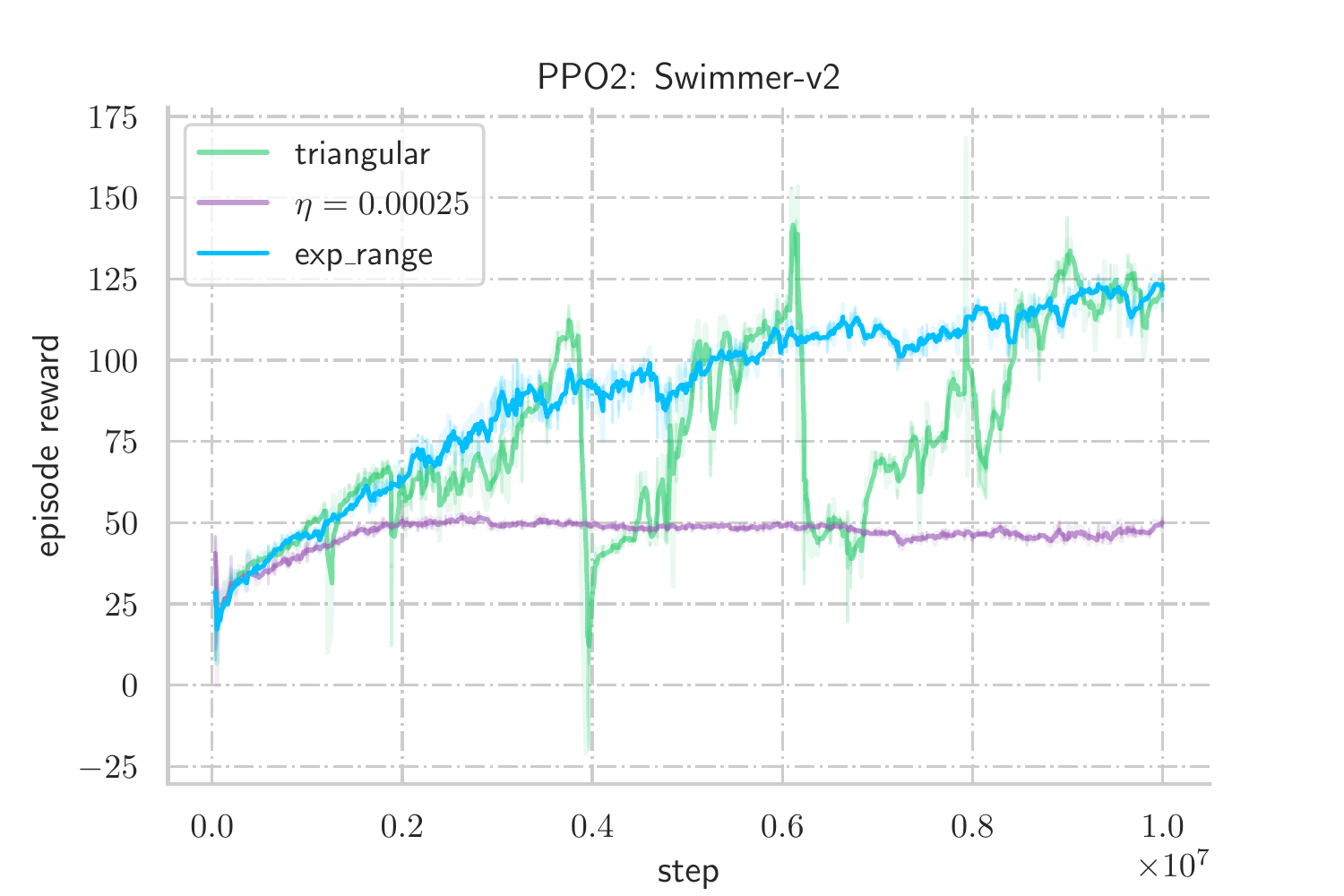}  
  \caption{}\label{fig:experiments:swimmer}
\end{subfigure}
\hfill
\begin{subfigure}{.49\textwidth}
  \centering
    \includegraphics[width=\linewidth]{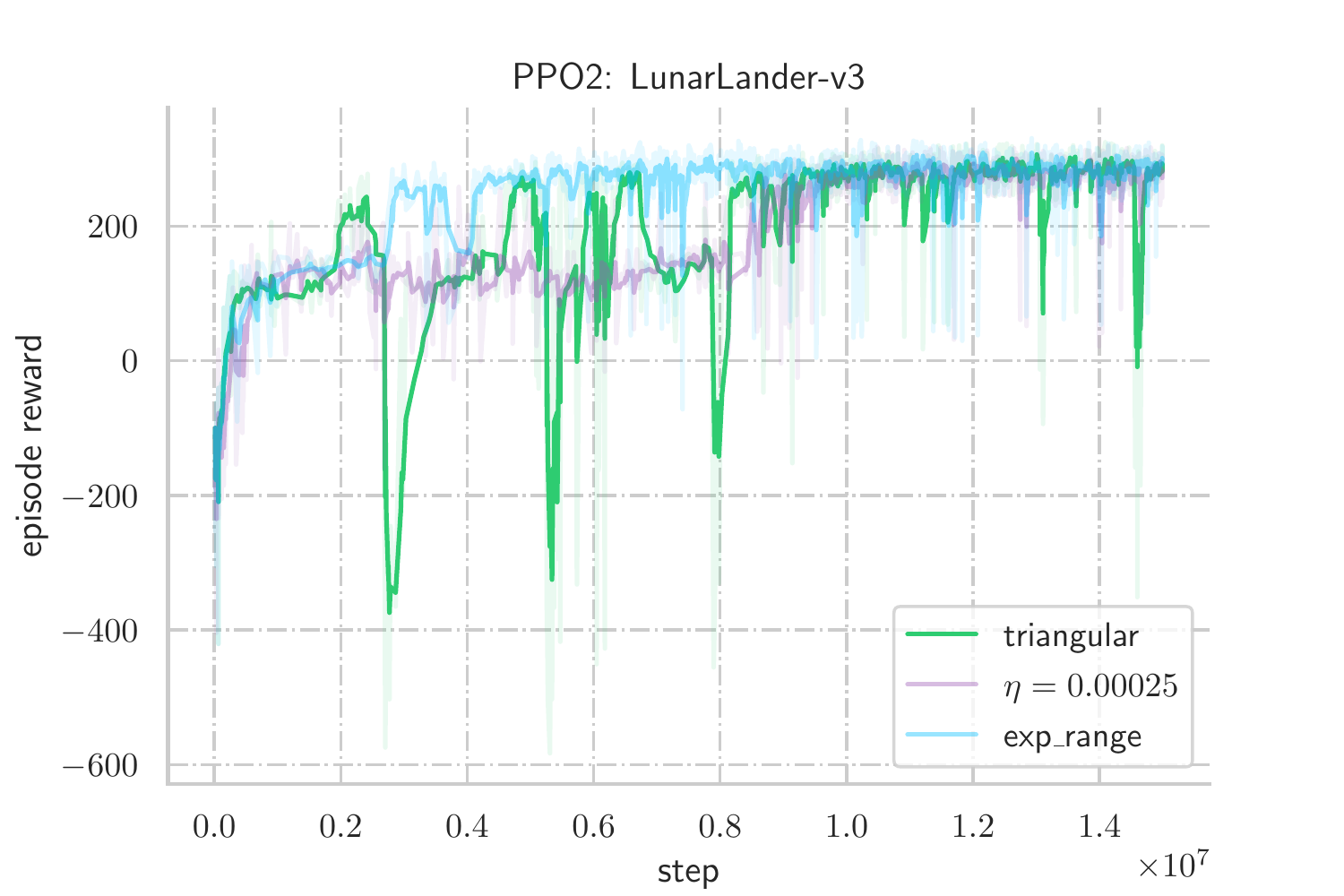}  
  \caption{}\label{fig:experiments:lunar}
\end{subfigure}
\caption{Results of training DRL agents using the PPO2 algorithm in various RL environments. Each subfigure illustrates the episode reward over time steps for three training processes on the same environment: applying the \textit{triangular} policy, the \textit{exp\_range} policy or using a tuned fixed learning rate.}\label{fig:experiments}
\end{figure*}

Results of the experiments are shown in figure~\ref{fig:experiments}.
Each plot contains the episode reward of the training process using the \textit{triangular} policy (labelled as \textcolor{flatgreen}{triangular}), the training process using the \textit{exp\_range} policy (labelled as \textcolor{deepskyblue}{exp\_range}) and the training process using the  optimal fixed learning rate (labelled with \textcolor{flatpurple}{$\eta = 0.001$}).

Figure~\ref{fig:experiments:cartpole} shows results of experiments on the \textit{CartPole-v0} environment.
Learning rate cycling introduces a longer period of exploration in the beginning of the training process but once converged, the episode reward does not drop in a later stage of training.
The \textit{exp\_range} policy reduces this effect substantially.
Note, that for experiments in the \textit{BipedalWalker-v3} environment (Fig.~\ref{fig:experiments:bipedal}) our method applies learning rates to the optimization process that significant higher (are two orders of magnitude) than the optimal learning rate of $\eta = 0.00025$.
This experiment also shows a longer period of exploration using learning rate cycling, which is consistent to the notion, that training at higher learning rates introduces regularization to the training process, e.g. no overfitting to a specific policy in the RL setting.
In particular, when using learning rate cycling, the agent is capable of achieving an overall higher reward compared to the optimal learning rate for that particular environment.
Again, the effects are reduced when applying the \textit{exp\_range} policy.
Experiments on the \textit{Swimmer-v2} environment (Fig.~\ref{fig:experiments:swimmer}) demonstrate that training with cyclical learning rates can achieve higher rewards than training with a fixed learning rate.
The \textit{triangular} policy results in a higher exploration while the \textit{exprange} policy produces a more robust training process.
Figure~\ref{fig:experiments:lunar} shows experiments on the \textit{LunarLander-v3} environment wherein all training processes achieve similar results.
However, the \textit{triangular} policy again imposes more exploration and the \textit{exp\_range} policy is capable of achieving the best results with the fastest convergence.
In this setting, the exponential decay is able to stabilize the learning process.
\section{Ablation Study}\label{sec:ablation}
Since general statements on methods in DRL and DL often do not hold when varying parameters, we perform an ablation study to identify critical parameters of our method. This helps to identify the most influential and essential components for the success or failure of training.

\textbf{Very High LR.} Very high learning rates lead to divergence.
Defining $\eta_{max} = 0.1$ or higher leads to divergence in each environment of the experiments.
This is due to the instability that high learning rates impose on the optimization process.

\textbf{Learning Rate Finding.} Learning rate finding is the procedure of training a model with a few batches, each training step with a different learning rate.
The optimal learning rate is the one at which the loss decreases the most.
In the supervised setting, appropriate learning rates per epoch can be identified by evaluating varying learning rates applied to the training process. However, due to the non-stationarity of the RL problem, this technique is not trivially transferable.
This is especially the case when using on-policy RL, where each update is performed using a new batch from the previous rollout. Transferring this learning rate finding technique and applying it to our training procedure leads to divergence for each experiment in each environment.

\textbf{Pendulum-v0.} Training an agent at cyclical learning rates in the \textit{Pendulum-v0} environment leads to divergences in each environment.
We attribute this to a poor combination of hyperparameters for this environment.

\section{Conclusion}\label{sec:conclusion}
We apply learning rate cycling, first introduced in~\cite{smithCyclicalLearningRates2017}, to DRL by training agents on various environments using the PPO2 algorithm with cyclical learning.
Experiments show that, training with cyclical learning rates is capable of developing strategies to achieve similar or better results than training with a fixed learning rate.
Most notably, our method is capable of achieving these results without manually tuning the learning rate bounds for  specific environments.
Hence, we take a step towards reducing the amount of hyperparameters to be manually tuned in Deep Reinforcement Learning (DRL) training processes. Whether this technique can successfully be applied to off-policy deep RL remains an open question.
This seems especially appealing since off-policy DRL training is known to be prone to divergence~\cite{sutton1998introduction}.

\bibliography{root}{}

\end{document}